\documentclass{article}

    \PassOptionsToPackage{numbers, compress}{natbib}

\usepackage[preprint]{neurips_2020}

\usepackage[utf8]{inputenc} %
\usepackage[T1]{fontenc}    %
\usepackage{url}            %
\usepackage{booktabs}       %
\usepackage{amsfonts}       %
\usepackage{nicefrac}       %
\usepackage{microtype}      %

\usepackage{color}
\usepackage{graphicx}
\usepackage{bbm}
\usepackage{amsmath}
\usepackage{floatrow}
\usepackage{subcaption}
\usepackage{makecell}
\usepackage{multirow}
\usepackage{enumitem}

\definecolor{citecolor}{RGB}{34, 139, 34}
\usepackage[pagebackref=true,breaklinks=true,colorlinks,bookmarks=false, citecolor=citecolor]{hyperref}

\title{Frustratingly Simple Domain Generalization via Image Stylization}

\author{%
  Nathan Somavarapu, Chih-Yao Ma, Zsolt Kira\\
  Georgia Institute of Technology\\
  Atlanta, GA 30332 \\
  \texttt{\{nsomavarapu3,cyma,zkira\}@gatech.edu} \\
}

\begin{document}
\maketitle

\begin{abstract}
  Convolutional Neural Networks (CNNs) show impressive performance in the standard classification setting where training and testing data are drawn i.i.d. from a given domain. However, CNNs do not readily generalize to new domains with different statistics, a setting that is simple for humans. In this work, we address the Domain Generalization problem, where the classifier must generalize to an unknown target domain. Inspired by recent works that have shown a difference in biases between CNNs and humans, we demonstrate an extremely simple yet effective method, namely correcting this bias by augmenting the dataset with stylized images. In contrast with existing stylization works, which use external data sources such as art, we further introduce a method that is entirely in-domain using no such extra sources of data. We provide a detailed analysis as to the mechanism by which the method works, verifying our claim that it changes the shape/texture bias, and demonstrate results surpassing or comparable to the state of the arts that utilize much more complex methods\footnote{Code will be released at \url{https://github.com/GT-RIPL/DomainGeneralization-Stylization}.}.
\end{abstract}
\section{Introduction}
\label{sec:intro}

Humans demonstrate an amazing ability to respond robustly to a number of different settings. 
For example, humans can generally deal with moderate changes in illumination, background or color when identifying an object. 
One of the central goals of vision research is to build methods which share this ability of humans. 
On the other hand, a significant amount of research in machine learning has gone into fully supervised methods under the i.i.d setting. 
Although this has led to remarkable results, supervised methods still demonstrate frailty. 
\citet{recht2019imagenet} show that computer vision models trained on ImageNet~\cite{deng2009imagenet} and CIFAR10~\cite{krizhevsky2014cifar} do not generalize to data collected through a nearly identical process.
Similarly, \citet{hendrycks2019benchmarking} show that modern convolutional neural network (CNN) architectures display poor robustness to common perturbations that humans have no problem dealing with. 
Partially accounting for this, the biases of CNNs are significantly different than that of humans, which leads to a discrepancy in how decisions are made \cite{geirhos2018imagenet}.

Domain Adaptation (DA) is an area of work that attempts to deal with this issue by making use of unlabeled target data along with data from one or more source domains. 
A number of distribution alignment methods have been proposed with good empirical performance~\cite{ganin2014unsupervised, shu2018dirt, tzeng2017adversarial, long2015learning,long2016unsupervised}. 
A more general version of this problem is Domain Generalization (DG) where no (even unlabeled) target data is available. 
This setting is scientifically interesting and very practically motivated, since ideally a model should not need to be retrained in new domains when deployed in the world.
Due to the lack of target domain knowledge, it is key to consider the biases of the model to ensure features are robust across domains are used.
A line of previous work has approached this through the use of distribution matching techniques across the source domains via adversarial distribution alignment with MMD~\cite{li2018domain} or by minimizing domain dissimilarity \cite{muandet2013domain}.

In this paper, in contrast to distribution matching, we explicitly consider model biases which are more generalizable by decreasing reliance on irrelevant features. Specifically, 
we hypothesize that implicit variability within the source domains themselves can be leveraged to train  models that focus more on robust features (e.g. shape), as opposed to less relevant features (\textit{e.g.,} textural information). Such models would likely be more robust when generalizing out of domain~\cite{mitchell1980need}. 
To this end, we first introduce a simple yet effective probabilistic data augmentation method which varies the texture of images
by stylizing them in the style of paintings.
Intuitively, by introducing a variance in the local information, while maintain the same class label, we hope that the model relies more heavily on the more general feature of shape instead of the less general feature of texture.
This shift in bias, as discussed previously, is crucial in the DG setting.
Combining this augmentation with batched source training, we show that this comparatively simple method surpasses state of the arts or achieves comparable performance on the common DG datasets of PACS~\cite{li2017deeper}, VLCS~\cite{fang2013unbiased} and Office-Home~\cite{venkateswara2017Deep}. 
We additionally introduce a (pairwise) Inter-Source stylization method, which uses \textit{only} in-domain information and eliminates the requirement of using an external source for stylization.
We also empirically show that this method consistently produces improvement across different backbone networks. 
Finally, we provide several quantitative analyses to show how the proposed method impacts the model's shape and texture bias. 
The analysis leads to several interesting findings, for example, (1) shape accuracy is increased but texture accuracy is largely unchanged,
(2) surprisingly, Inter-Source stylization achieves similar performance to stylization with external painting dataset, indicating that there is enough cross-source variability to leverage through stylization.

Our contributions are summarized below:
\begin{itemize}[topsep=0pt,itemsep=-1ex,partopsep=1ex,parsep=1ex,labelindent=0.0em,labelsep=0.2cm,leftmargin=*]
    \item We introduce a simple transformation based on stylization which uses only in-domain data and surpasses or is competitive with current state of the arts for domain generalization.
    \item We provide insights into the inner workings of our method by generating a cue-conflict dataset and comparing the shape bias, shape accuracy, and texture accuracy of models trained in different ways.
    \item We observed that a model trained with stylization increases its predictiveness of shape but interestingly still maintains predictiveness of texture.
\end{itemize}

\section{Related Work}
\label{sec:related}

\subsection{Domain Generalization}
The goal of domain generalization (DG) is to learn a representation which generalizes over multiple \textit{unknown} data distributions. 
DG captures real world scenarios by providing a number of source domains that can be used for training and an unseen target domain that the model is tested on. 
A large amount of previous work in this area relies on methods that design models which reduce reliance on domain specific artifacts. 
Some of these methods include low-rank parameterized models~\cite{li2017deeper}, domain specific classifiers which are fused by an aggregation module~\cite{d2019domain}, unsupervised latent discovery of domains~\cite{matsuura2019domain} and meta-learning~\cite{dou2019domain, yu2018one}. 
Other methods rely on augmentation of source data with the hope of capturing the target domain in either image space~\cite{tobin2017domain} or in an embedding space~\cite{volpi2018generalizing}. 
Still other methods employ self-supervised losses which are paired with the standard classification loss, such as \cite{carlucci2019domain} which uses the jigsaw puzzle task.
Our method just relies on a simple augmentation during source training, but focuses on a specific form of augmentation (stylization) that can use \textit{inter-source information} in order to implicitly improve invariance to irrelevant features. We note that this method is extremely simple, compared to more sophisticated methods utilizing meta-learning, classifiers for specific domains, unsupervised domain discovery, etc.

\subsection{Network Bias}
Since in DG there is no target data to rely on, we propose that 
the implicit biases in the model are important since they must tend towards those that are more generalizable to unknown domains, even though only source domains are available. 
A large body of work has shown that CNNs likely rely on significantly different features than humans for classification. 
Early on \citet{gatys2015texture}, while designing a method for texture synthesis, showed that a linear classifier on top of the texture representations of VGG19 \cite{simonyan2014very} achieves nearly the same classification accuracy as a VGG19 network directly trained on the task.
Later \citet{jo2017measuring} showed that networks latch on to statistical regularities in the Fourier statistics which are independent of the actual content of the image.

A line of recent work has started to consider the biases of CNNs and effects that this has, \citet{wang2019learning} seek to penalize the predictive power of the early layers of the network via an adversarial loss. 
\citet{geirhos2018imagenet} showed that, while humans mostly rely on shape for classification, common CNNs rely far more on texture and low-level information.
They introduce \textit{stylization} to study the comparative reliance of humans and CNNs on shape and texture. Specifically, stylization (or style transfer) methods seek to match the content of a particular image to the style of a separate image. 
The pioneering work of \citet{gatys2015neural} uses the same Gram Matrix based texture representation from ~\citet{gatys2015texture} paired with activations from different layers of a CNN to match the style and content of two images through an iterative optimization process. 
Using this idea, \citet{geirhos2018imagenet} introduce stylized ImageNet, a dataset with ImageNet \cite{deng2009imagenet} images in the style of common paintings.
A model is then trained on this dataset and fine-tuned on ImageNet to achieve improved ImageNet validation performance. In addition it is shown that this process increases the shape bias of the model.

In this paper, we extend these works by first introducing stylization as a probabilistic augmentation. 
Existing works have also used stylization for augmentation~\cite{jackson2018style} but have not focused on domain generalization, and the biases that are introduced as a result, and have also used external sources as the style sources hence using extra data.
Importantly, we show that the same results can be attained by only using \textit{inter-source stylization} in the domain generalization setting, \textit{i.e.,} we do not rely on the addition of the extra painting dataset for a source of style (see Sec.~\ref{sec:experiments}). 
We hypothesize that variability in the sources allows the model to learn representations that are more robust to irrelevant features, specifically increasing its shape bias and reducing its texture bias, leading to improved generalization. We further provide extensive analysis showing this, specifically in the context of shape/texture bias. 
\begin{figure}[t]
    \begin{center}
    \includegraphics[width=0.85\linewidth]{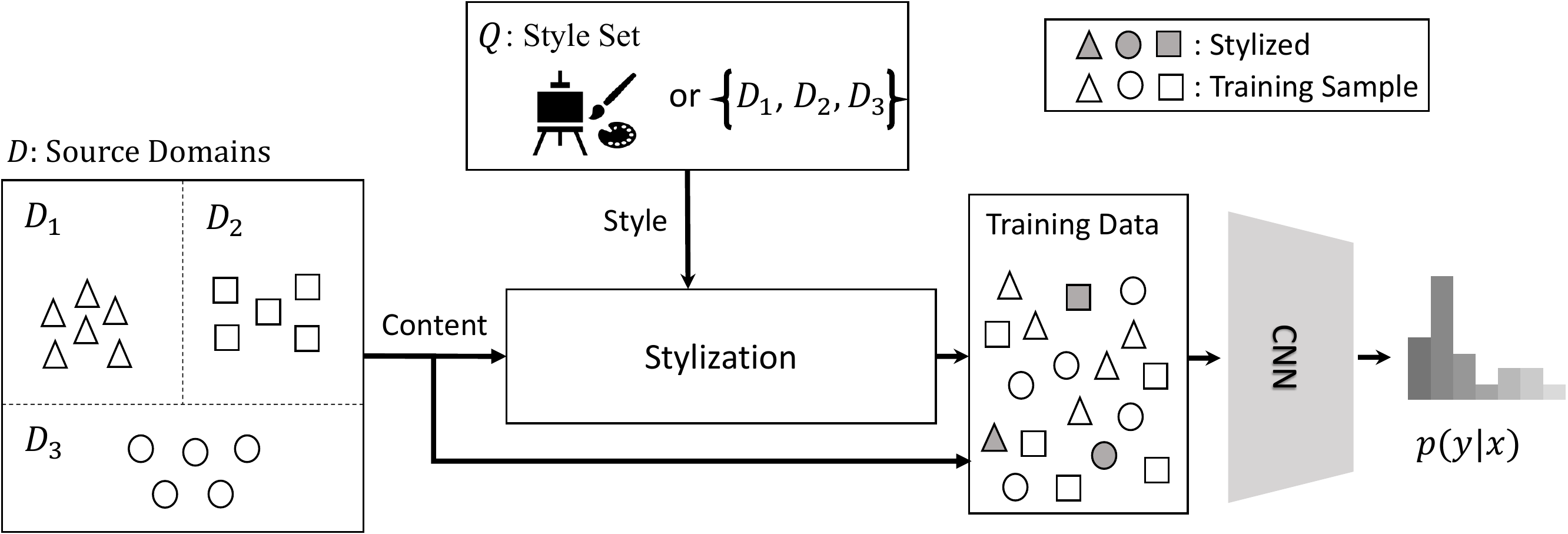}
    \caption{
        Illustration of  training pipeline of the Proposed Method. Each image from the source domain is replaced with a stylized version of itself with a certain probability. Paintings are used to stylize in the standard setting; however, to avoid the introduction of extra datasets, the \textit{source} domains themselves can be used as sources of style. 
    }
    \label{fig:concept}
    \end{center}
\end{figure}

\section{Image Stylization for Domain Generalization}
\label{sec:method}

In this paper, we tackle the Domain Generalization (DG) problem, where there are multiple source datasets and the goal is to learn models generalizable to unknown target datasets, where there is some domain shift. We denote the source distributions $\{ D_{S_{1}}, ..., D_{S_{n}} \}$ and the target distribution $D_{T}$. 
Note that in most of the DG datasets $n=3$.
With a slight abuse of notation we denote samples drawn randomly from one of the source domains as $x,y \sim \{ D_{S_{1}}, ..., D_{S_{n}} \}$. 
We also use $f(x;\theta): X \rightarrow \mathbb{R}^{c}$ to denote a CNN with softmax output, where $c$ denotes the number of classes. Finally we denote the standard cross entropy classification loss as $\mathcal{L}_{c}(f(x;\theta), y)$.

The key intuition for our method stems from the perspective of increasing the model bias towards features that are more generalizable, by decreasing reliance on irrelevant features even if they perform well on the (validation set of) source domains. 
Specifically, we hypothesize that having access to multiple source domains or even sufficient variation within a single domain can be leveraged in order to infer types of variability irrelevant to generalization (\textit{e.g.,} texture bias) to further encourage relevant features (\textit{e.g.,} shape bias). 
Similar to data augmentation methods~\cite{volpi2018generalizing, choi2019self, huang2018auggan}, we can use this to learn models robust to such variations, but in a manner that does not requiring specifying all of the possible transformations. 
In summary, we seek to expose the network to a variety of irrelevant features such as varying textures, which is implicitly available in the multi-source data, so as to make the network more robust when applied out of domain. 
A key question is how to leverage such inter-source variability implicitly available in the data. We propose that \textit{stylization} is one such method to do this; since each domain contains its own set of irrelevant features, transforming the content of one source domain using the style of other source domains presents a simple method for utilizing multiple variations implicitly available in the data and which do not destroy the semantic meaning (since content is preserved).

In the next section, we introduce probabilistic stylization in order to implement this idea, and show that unlike traditional stylization (which uses external data sources such as art) we can apply \textit{inter-source stylization} to the same effect as well.

\subsection{Probabilistic Stylization}

\begin{figure}[t]
    \begin{center}
    \includegraphics[width=0.90\linewidth]{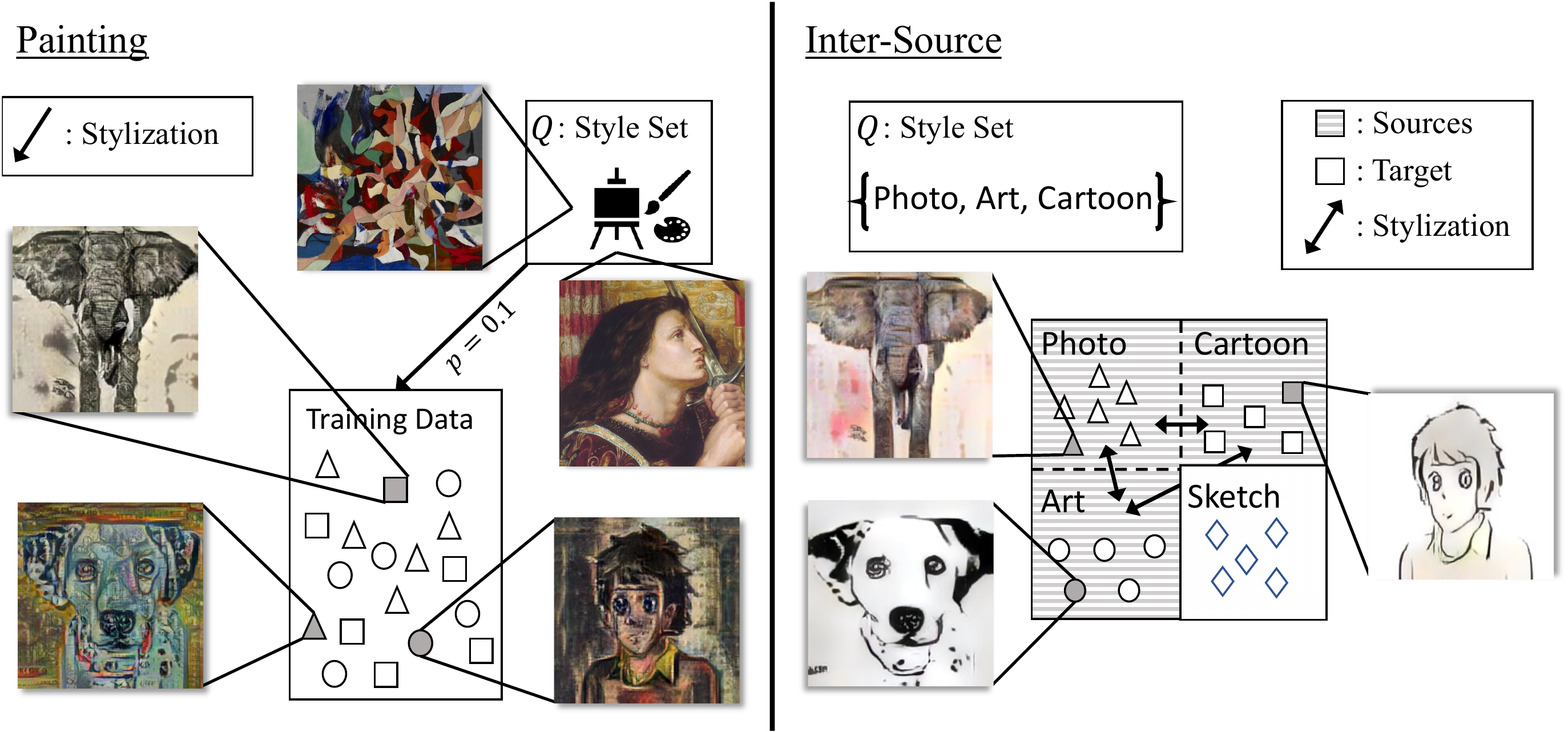}
    \caption{
        Samples from painting stylized PACS images and Inter-Source Stylized PACS images. On the right, the inter-source stylized images depict when one domain is used as the content with another used as the style.}
    \label{fig:stylized_sample}
    \end{center}
\end{figure}

Here we define a simple transformation which we denote $S_{Q,p}(x): \mathbb{R}^{NxN} \rightarrow \mathbb{R}^{NxN}$. 
This transformation replaces the original image $x$ with a stylized version of itself with probability $p$, where the style is drawn uniformly at random from the style set $Q$. 
The transformation is implemented using AdaIN~\cite{huang2017arbitrary} which employs a stylization network to achieve fast stylization to arbitrary styles.
In the initial setting, we follow many of the stylization works and select the set $Q$ to be the painter-by-numbers dataset\footnote{\href{https://www.kaggle.com/c/painter-by-numbers}{https://www.kaggle.com/c/painter-by-numbers}}. 
Some examples of this stylization for content images from the PACS dataset can be seen in Figure \ref{fig:stylized_sample}. We subsequently relax this to not include external sources.

Our entire training process can be described with the following equation (depicted in Figure~\ref{fig:concept}). Intuitively, this shows the standard classification training process with random application of stylization.
\begin{equation}
    \min_{\theta}  \mathop{\mathbb{E}_{x,y \sim D_{S}}\mathcal{L}_{c}(f\left( S_{Q,p}(x);\theta), y \right)}
\end{equation}

\subsection{Inter-Source Stylization}
Leveraging the painter-by-numbers dataset as the source of stylization
introduces an additional source of information into the training process even if does not contain labels, making it difficult to compare under the standard DG setting. 
To overcome this, we propose a method which simply uses the other available sources for stylization. 
Surprisingly, as described in Section 5, this achieves similar performance to stylization with the painting dataset, indicating that there is enough cross-source variability to leverage through stylization. 
This amounts to modifying the set $Q$ in our stylization transformation, $Q = \{ D_{S_{1}}, ..., D_{S_{n}} \} / D_{c}$, where $D_{c}$ denotes the domain of the image currently under stylization.
Taking this to the extreme we also experiment with intra-source stylization where stylization happens only \textit{within} a single domain, i.e. $Q = D_{c}$.
For example in this setting a image from the photo domain could only be stylized as other photo images.
Some examples of PACS content images generated through the inter-source stylization process can be seen in Figure 2.
Images stylized through inter-source stylization have more visual variability than those stylized solely to paintings, however, both retain important semantic attributes.
The training process for inter-source stylization then becomes optimizing the following objective:
\begin{equation}
    \min_{\theta} \mathop{\mathbb{E}_{x,y \sim \{ D_{S_{1}}, ..., D_{S_{n}} \} / D_{c} }\left[\mathcal{L}_{c}(f\left( S_{D_{S},p}(x);\theta), y \right)\right]}
\end{equation}

\section{Experiments}
\label{sec:experiments}

We benchmark our method over a number of common DG datasets: PACS \cite{li2017deeper}, VLCS~\cite{fang2013unbiased}, and Office-Home \cite{venkateswara2017Deep} datasets. 
First, we validate the proposed method on the PACS dataset which spans four distinct categories: \textit{Photo}, \textit{Art}, \textit{Cartoon} and \textit{Sketch}. 
We then evaluate our method on the VLCS~\cite{fang2013unbiased} dataset which contains only photos, but from different datasets. 
The PACS dataset contains 9991 images while the VLCS dataset contains 10,729 images.
Both the PACS and VLCS datasets contain a small number of classes, with 7 and 5 respectively.
Thus, to demonstrate the effectiveness of our method on relatively larger datasets, we test on the Office-Home dataset which contains 65 classes and 15,500 images.
In each setting, we compare with the state of the arts, and our method significantly surpasses or is comparable to much more complex methods. 
We further provide some analysis that indicates that the network became more shape accurate, slightly less texture accurate and became more biased to shape through this process; see Sec.~\ref{sec:analysis} for more detail.

\subsection{PACS Dataset}
We adopt the testing procedure detailed by \citet{li2017deeper}, for the four domains \textit{Photo}, \textit{Art}, \textit{Cartoon} and \textit{Sketch}.
In this setting, three domains are used for training and the remaining domain is the testing domain. Results are provided for each testing domain and  averaged over all testing domains. 
Each dataset contains the same seven classes: \textit{dog}, \textit{elephant}, \textit{giraffe}, \textit{guitar}, \textit{horse}, \textit{house}, and \textit{person}.

\textbf{Training detail.}
Optimization is performed via Stochastic Gradient Descent (SGD) with a learning rate of 0.001 and momentum of 0.9, a weight decay of 0.0005, and batch size of 128. 
The model is trained for 80 epochs with a reduction in the learning rate at epoch 60. 
The stylization probability and stylizing strength is set to $10\%$ and $100\%$, respectively. 
Similar to prior work we employ common data augmentations of horizontal flipping with $50\%$ probability, random cropping retaining $80\% - 100\%$ of the image and random color jitter. 
The hyper-parameters and the model evaluated on the test domains are chosen via the \textit{source} validation set accuracy of all sources batched together averaged over all domains.
We employ Resnet18 and AlexNet models pre-trained on ImageNet for all experiments, following the methodology introduced in~\citet{li2017deeper} and used in contemporary work.

We report average accuracy over three independent runs of the method in Table~\ref{PACS-table}\footnote{\label{clarity}For visual clarity an identical table with standard deviation can be found in the supplementary materials.}.
We can see that our method achieves state of the art results in all four domains and it is interesting to note that our method has the largest performance gain in the \textit{Sketch} setting.
Here shape information is the most important, as features based on local textural information will likely not generalize.
In Sec.~\ref{sec:analysis}, we provide further analysis as to the relationship of model invariances and performance in the \textit{Sketch} domain supporting this view of generalization.
We can also see that the differences between stylization within domain and stylization with the painter-by-numbers dataset is negligible.
On the other hand, lack of a large amount of inter-source variability affects the performance of intra-source stylization, although it still contains enough variability to significantly benefit generalization.
The method is applicable across different models as can be seen from Table~\ref{PACS-table}. Both AlexNet and ResNet18 benefit from the introduction of stylized augmentation indicating that benefits of proper model bias are not model specific.

\begin{table}[t]
  \caption{
  Comparison with state of the arts on the PACS~\cite{li2017deeper} Dataset.
  Asterisk(*) denotes a different training/testing procedure, which computes max target dataset accuracy over the training period in lieu of using in source validation data for model selection.
  }
  \label{PACS-table}
  \centering
  \resizebox{\textwidth}{!}{
  \begin{tabular}{lllll|l||llll|l}
    \toprule
     & \multicolumn{5}{c}{ResNet-18} & \multicolumn{5}{c}{AlexNet} \\
    \midrule
    Method              & Photo & Art   & Cartoon   & Sketch    & Avg.  & Photo & Art   & Cartoon   & Sketch    & Avg.   \\
    \midrule
    D-SAM~\cite{d2019domain}       & 95.30 & 77.33 & 72.43     & 77.83     & 80.72 & 85.55 & 63.87 & 70.70     & 66.66     & 71.20\\
    MMLD~\cite{matsuura2019domain}        & 96.09 & 81.28 & 77.16     & 72.29     & 81.83 & 89.00 & 69.27 & \bf{72.83}     & 66.44     & 73.38\\
    MLDG~\cite{li2018learning}        &  -      & -     &  -    &  -        &  -        & 88.00 & 66.23 & 66.88     & 58.96     & 70.01\\
    Meta-Reg~\cite{balaji2018metareg}            & 95.50 & 83.70 & 77.20     & 70.30     & 81.70 & \bf{91.07} & 69.82 & 70.35     & 59.26     & 72.62\\
    Epi-FCR~\cite{li2019episodic}             & 93.90 & 82.10 & 77.00     & 73.00     & 81.50 & 86.10 & 64.70 & 72.30     & 65.00     & 72.00\\
    JiGen~\cite{carlucci2019domain}               & 96.03 & 79.42 & 75.25     & 71.35     & 80.51 & 88.98 & 67.63 & 71.71     & 65.18     & 74.38\\
    PAR~\cite{wang2019learning}       &  -      & -     &  -    &  -        &  -         & 90.40 & 68.70 & 70.50     & 64.60     & 73.54\\
    Ours~(Painting)     & 96.29 & \bf{84.60}  & 78.81    & 78.90      & \bf{84.64} & 90.00  & \bf{73.71}  & 72.37     & 70.40 & \bf{76.62} \\
    Ours~(Intra-Source) & \bf{96.45}  & 80.81  & 78.85      & 77.83 & 83.49 & 90.72  & 71.09  & 73.14   & 68.67    & 75.90 \\
    Ours~(Inter-Source) & 96.27  & 83.17  & \bf{78.90}      & \bf{79.87} & 84.55 & 89.76  & 72.52  & 71.72   & \bf{71.21}    & 76.30 \\
    \midrule
    MASF~\cite{dou2019domain}*                   & 94.99 & 80.29 & 77.17 & 71.69     & 81.79 & 90.68 & 70.35 & 72.46 & 67.33     & 75.21\\
    Ours~(Painting)*        & 96.85 & \bf{85.45}  & 79.56  & 79.32      & 85.30  & \bf{91.60}  & \bf{76.01}  & \bf{73.82}  & 73.07    & \bf{78.62} \\
    Ours~(Intra-Source) & \bf{97.05}  & 82.73  & 79.64      & 80.10 & 84.88 & 91.49  & 73.06  & 73.74   & 72.33    & 77.66 \\
    Ours~(Inter-Source)*    & 96.89  & 84.33 & \bf{80.15} & \bf{81.53}    &\bf{85.72}  & 91.52  & 74.59  & 73.15  & \bf{74.71}      & 78.49 \\
    \bottomrule
  \end{tabular}
  }
\end{table}

\subsection{VLCS Dataset}
VLCS is another widely used
DG dataset which is collected from four classic datasets, PASCAL (VOC)~\cite{everingham2010pascal}, LabelME~\cite{russell2008labelme}, Caltech~\cite{fei2004learning}, and SUN~\cite{choi2010exploiting}. 
Each domain in this setting contains the five classes \textit{bird}, \textit{chair}, \textit{car}, \textit{dog}, \textit{person} and as before three domains are chosen for training with the held out domain used for testing.
In this setting, we use AlexNet~\cite{krizhevsky2012imagenet} for comparison purposes to prior work. 
The same training parameters are used as in the PACS setting except a learning rate of 0.0005 and a weight decay of 0.00005, each of which was tuned on the source validation set.
In Table~\ref{VLCS-table}\textsuperscript{\ref{clarity}}, we show the performance of the stylization training method and compare with a number of different methods.
Our method performs comparably or exceeds the state of arts showing that shape information is helpful in this setting as well.
We note that although our method results in similar performance, methods like MMLD are significantly more complicated requiring a careful mix of clustering, adversarial loss, an entropy loss and a standard classification loss.

\begin{table}
\RawFloats
\begin{minipage}{.48\linewidth}
    \centering
    \resizebox{\textwidth}{!}{
    \begin{tabular}{lllll|l}
    \toprule
    \multicolumn{6}{c}{AlexNet}\\
    \midrule
                        & Pascal   & LabelMe   & Caltech   & Sun    & Avg.\\
    \midrule
    D-SAM~\cite{d2019domain}            & 58.
    59         & 56.95     & 91.75     & 60.84                      & 67.03 \\
    JiGen~\cite{carlucci2019domain}               & 70.62         & 60.90     & \bf{96.93}     & 64.30                      & 73.19 \\
    MMLD~\cite{matsuura2019domain}                & \bf{71.96}         & 58.77     & 96.66     & 68.13                      & 73.88 \\
    Epi-FCR~\cite{li2019episodic}            & 67.10         & \bf{64.30}     & 94.10     & 65.90                      & 72.90 \\
    MMD-AAE~\cite{li2018domain}             & 67.70         & 62.60     & 94.90     & 64.90                      & 72.28 \\
    Ours~(Painting)     & 69.81         & 58.90       & 96.57      & \bf{69.82}                     & 73.77  \\
    Ours~(Intra-Source) & 70.98  & 59.98  & 96.10  & 69.38                       & \bf{74.11}  \\
    Ours~(Inter-Source) & 71.17  & 60.19  & 96.27  & 67.68                       & 73.83  \\
    \midrule
    MASF*~\cite{dou2019domain}                  & 69.14 & \bf{64.90} & 94.78 & 67.64     & 74.11\\
    Ours~(Painting)*        & 72.04  & 61.97  & \bf{97.81}  & 70.67     & 75.62 \\
    Ours~(Intra-Source)*    & 72.28  & 61.57  & 97.51  & 70.76      & 75.53 \\
    Ours~(Inter-Source)*    & \bf{72.30}  & 61.95  & 97.51  & \bf{70.93}      & \bf{75.67} \\
    \bottomrule
  \end{tabular}
  }
    \caption{
    Comparison with state of the arts on VLCS~\cite{fang2013unbiased}.
    Asterisk(*) denotes a different training/testing procedure, same as in Table~\ref{PACS-table}.
    }
    \label{VLCS-table}
  \end{minipage}%
  \hspace{0.5cm}
  \begin{minipage}{.48\linewidth}
    \centering
    \resizebox{1\textwidth}{!}{
    \begin{tabular}{lllll|l}
    \toprule
    \multicolumn{6}{c}{ResNet-18} \\
    \midrule
                        & Art & Clipart & Product   & Real    & Avg. \\
    \midrule
    D-SAM~\cite{d2019domain}               & 58.03 & 44.37 & 69.22     & 71.45     & 60.77\\
    JiGen~\cite{carlucci2019domain}               & 53.04 & 47.51 & 71.47     & 72.79     & 61.20\\
    Ours~(Painting)     & \bf{59.87} & 49.91 & \bf{72.82}     & \bf{75.20}     & \bf{64.45}\\
    Ours~(Intra-Source) & 59.34 & 50.31 & 72.67     & 74.93     & 64.31\\
    Ours~(Inter-Source) & 59.52 & \bf{50.86} & 72.15     & 75.13     & 64.42\\
    \bottomrule
  \end{tabular}
  }
    \caption{
    Comparison with state of the arts on the OH \cite{venkateswara2017Deep} Dataset.
    }
    \label{OH-table}
\end{minipage}
\end{table}

\subsection{Office-Home Dataset}

The Office-Home Dataset \cite{venkateswara2017Deep} also contains four domains, \textit{Art}, \textit{Clipart}, \textit{Product}, and \textit{Real-World}. 
This dataset contains significantly more classes than either the PACS or VLCS datasets with 65 common items from the office and home. 
As employed in previous work, we adopt the testing procedure employed in~\cite{d2019domain} which is very similar to the PACS testing setting.
As before, three domains are used for training and the remaining domain is the testing domain, 90\% of the data is used for training with the remaining 10\% used for validation. 
In this setting we employ the same hyper-parameters used in the PACS setting with the ResNet18 model.
We can see from Table \ref{OH-table}\textsuperscript{\ref{clarity}} that, even with a significant increase in classes, stylization is able to help the model generalize, validating the efficacy of shape bias.  

\begin{figure}[t]
    \begin{center}
    \includegraphics[width=1\linewidth]{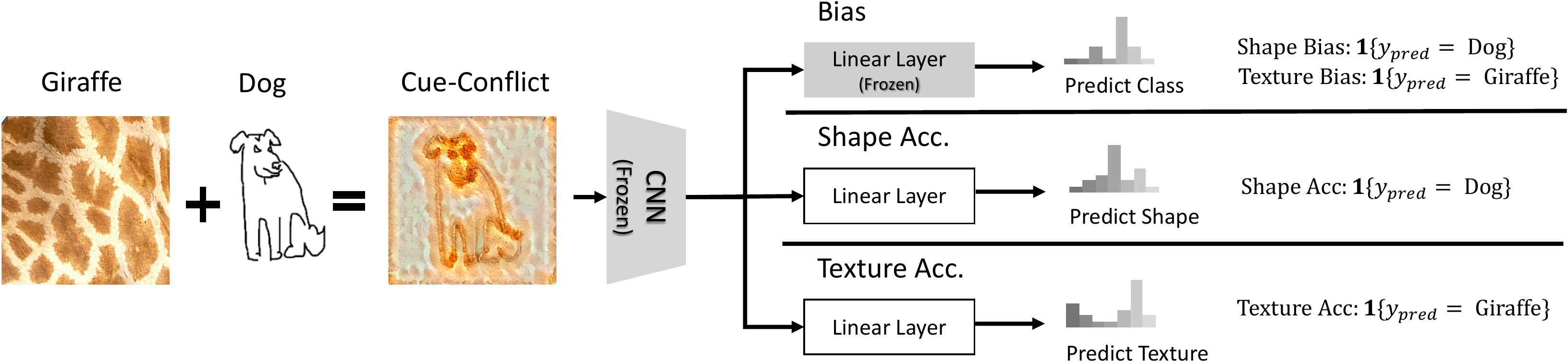}
    \caption{
    Visual depiction of shape bias, shape accuracy and texture accuracy.
    }
    \label{fig:shape_bias}
    \end{center}
\end{figure}

\section{Analysis}\label{sec:analysis}

 We now analyze various aspects of our approach, including an analysis of shape/texture predictiveness and bias, as well as analysis of which sources are most useful for use in stylization. 
 
\subsection{How does our method affect shape bias, shape accuracy, and texture accuracy?}

Here we provide some experimentation which sheds some light on how our method works. 
Inspired by \cite{geirhos2018imagenet} and \cite{hermann2019exploring}, we seek to measure the difference in shape bias, shape accuracy and texture accuracy of the model in order to test our hypothesis that our method decreases reliance on less generalizable features (texture) and increases reliance on more generalizable features (shape).

\textbf{Dataset preparation.}
In order to compute each of the above metrics, we first generate a cue-conflict dataset, in which the shape of the object in the image conflicts with the texture of the images (see Figure \ref{fig:shape_bias}). 
These images are generated by applying the stylization method of \cite{gatys2015texture} to a texture and content image. 
A set of textures were sourced from online in a manner identical to \cite{geirhos2018imagenet} and for content images we used images drawn from the Sketch dataset in PACS. 
Each texture and content image is chosen so that all models under comparison predict the correct class.
For example, a dog texture image is only taken if every model correctly predicts it as a dog, and a sketch of a house is only chosen as a content image if every model correctly predicts it as a house.
This limitation results in a maximum of 45 images per a class in the dataset as this is the maximal number of sketch images classified correctly by the least performant model we seek to analyze.
Finally, to actually generate the dataset, each content image was stylized with a randomly chosen texture image, excluding the setting where the content class and texture class match.
We chose this setting since the Sketch dataset is the most shape reliant dataset and since this presents the largest increase in accuracy in the PACS dataset (Table~\ref{PACS-table})
Also silhouettes were employed in~\cite{geirhos2018imagenet}, as disambiguation of shape and texture effects requires a silhouette, with no added texture, for the shape class.
In addition, the selection of arbitrary texture and shape not seen in the dataset is important analysis for the DG setting since we wish to see how a model performs when it is not trained on any of the target data.

\textbf{Shape bias.}
Shape bias is defined for a specific class as the proportion of class predictions which the model makes based on shape, in the cue-conflict dataset. 
We can define a sample from the cue-conflict dataset as $(x, y_{s}, y_{t})$, where $x$ is the cue-conflict image and $y_{s}, y_{t}$ represent the shape label and texture label respectively. 
The shape bias for a particular class $c$ is then the percentage of the time that the model predict the shape class for a cue conflict image when it predicts class $c$. Formally, it can be defined as follows:
\begin{equation}
    \frac{\sum_{i} \left(\mathbbm{1}[f(x^{i};\theta) = y^{i}_{s}] \ \text{and} \ \mathbbm{1}[f(x^{i};\theta) = c] \right)}{\sum_{i} \left(\mathbbm{1}[f(x^{i};\theta) = y^{i}_{s}] \ \text{and} \ \mathbbm{1}[f(x^{i};\theta) = c] \right) + \sum_{i} \left(\mathbbm{1}[f(x^{i};\theta) = y^{i}_{t}] \ \text{and} \ \mathbbm{1}[f(x^{i};\theta) = c] \right)}
\end{equation}

\textbf{Shape accuracy.}
To calculate shape accuracy, we freeze the weights of the model and attach a linear classifier to the penultimate layer of the ResNet18 model. 
We then train the classifier to predict the shape class of each image. 
Intuitively this metric measures the amount of shape information that the model contains. 
This is different than the shape bias since the bias measures the preference of the model, but does not preclude the existence of information in the model which it does not make use of.
Due to the small size of the cue-conflict dataset, following the procedure of \cite{hermann2019exploring}, we perform five-fold cross validation and report the average of the max validation accuracy over the training period over all five folds.

\textbf{Texture accuracy.}
Similar to shape accuracy, texture accuracy is computed by adding a linear classifier onto the penultimate layer of the ResNet18 model. 
Instead of training the model to predict the shape class, however, we train the model to predict the texture class. Here we hope to understand how much texture information is available in the model.

\begin{table}
\RawFloats
\begin{minipage}{.58\linewidth}
    \centering
    \resizebox{\textwidth}{!}{
    \begin{tabular}{lllllllll}
    \toprule
    \multicolumn{9}{c}{Shape Bias} \\
    \midrule
    & Dog & Elephant   & Giraffe   & Guitar & Horse & House & Person  & Avg. \\
    \midrule
    Basic                   & 0.00   & 10.64 & 2.17  & 10.91 & 0.00 & 0.00   & 12.50 &   5.17 \\
    Color Jitter            & 2.56   & 17.65 & 0.00  & 15.25 & 9.09 & 3.33   & 12.20 &   8.58 \\
    \multirow{2}{*}{\makecell{Style + \\ Color Jitter}}    & \multirow{2}{*}{\bf{15.79}}   & \multirow{2}{*}{\bf{41.67}} & \multirow{2}{*}{\bf{10.26}}  & \multirow{2}{*}{\bf{31.75}} & \multirow{2}{*}{\bf{24.24}} & \multirow{2}{*}{\bf{50.00}}   & \multirow{2}{*}{\bf{30.30}} &   \multirow{2}{*}{\bf{29.14}} \\ \\
    \bottomrule
  \end{tabular}
  }
    \caption{Shape Bias of models trained with style and color jitter, color jitter, no augmentation.}
    \label{bias}
  \end{minipage}%
  \hspace{0.2cm}
  \begin{minipage}{.4\linewidth}
    \centering
    \resizebox{1\textwidth}{!}{
    \begin{tabular}{llll}
    \toprule
    & Shape & Texture  & PACS Sketch \\
    \midrule
    Random Untrained        & 26.35  & 63.81 & 2.04\\
    Basic                   & 46.99  & \bf{100.00} & 65.66\\
    Color Jitter            & 52.70  & 99.36 & 71.91\\
    Style + Color Jitter    & \bf{59.05}  & 99.05 & \bf{78.90} \\
    \bottomrule
  \end{tabular}
  }
    \caption{
  Shape Accuracy and Texture Accuracy of models trained with style and color jitter, color jitter, no augmentation and a random untrained model}
    \label{shape-texture-acc}
\end{minipage}
\end{table}

We compare three models, a model trained with no augmentation, a model trained with augmentation from prior work consisting of horizontal flipping, color jitter and random crops and a model trained with our stylization method combined with basic augmentation. 
We perform this experiment on each of these models so that the progression of shape importance is clear. 
It is clear from Table \ref{bias} and Table \ref{shape-texture-acc} that the stylized training results in a model that has significantly higher shape bias and shape accuracy. 
Interestingly, however, the texture accuracy decreases only marginally. 
This is surprising since one might expect the introduction of different textures to encourage the network to become invariant to texture. 
One possible explanation for this could be that the classification layer of the model learns to ignore the textural information even though it may be present. 
An interesting finding is that as shape bias increases, the resulting domain generalization performance seems to increase as well; this validates the hypothesis that encouraging the correct biases can improve generalization. 

\subsection{Are all sources equally useful?}

Since we utilize inter-source variability through stylization, one can ask whether all sources are necessary or whether they equally benefit performance.
In Table~\ref{style-abl}, we show the effects of using different domains for inter-source stylization so as to ascertain which source domains are most important for each target dataset. 
The results show that not all source datasets are necessary, with one or two sources being important for each target. Further, different sources are important for different targets but in most cases using more than one source increases performance. For all cases, the performance of using all three sources is close the best condition implying that there is not a need to select among the sources. 

\begin{table}[t]
  \caption{Test Accuracy when only limited sources are available for stylization. The columns represent the sources that are used for inter-source stylization. The rows are the test dataset being tested on. Each result shown is the average over 3 independent runs.}
  \label{style-abl}
  \centering
  \resizebox{0.60\textwidth}{!}{
  \begin{tabular}{llllllll}
    \toprule
    & \multicolumn{7}{c}{Source Stylization Datasets} \\
    \toprule
    & A,C,S & A,S & A,C & C,S & A & C & S \\
    \midrule
    Photo~(P)   & 96.27 & 96.33 & 96.13 & 96.41 & 96.35 & \bf{96.49} & 96.19 \\
    \midrule
    \midrule
    & P,C,S & C,S & C,P & P,S & C & P & S \\
    \midrule
    Art~(A)   & 83.17 & 78.81 & 82.30 & 82.96 & 78.22 & \bf{83.32} & 79.75 \\
    \midrule
    \midrule
    & A,P,S & A,S & A,P & P,S & A & P & S \\
    \midrule
    Cartoon~(C)   & 78.90 & 78.56 & \bf{78.92} & 78.54 & 78.50 & 78.84 & 77.59 \\
    \midrule
    \midrule
    & A,P,C & A,C & A,P & P,C & A & P & C \\
    \midrule
    Sketch~(S)   & 79.87 & 80.27 & 75.40 & \bf{80.67} & 74.47 & 73.99 & 79.63 \\
    \bottomrule
  \end{tabular}
  }
\end{table}
\section{Conclusion}

In this paper, we introduced a simple technique, stylization, for improving out-of-domain performance by encouraging the model to learn more generalizable biases, \textit{e.g.,} shape.
Interestingly, we show that different domains contain variable information, which can be used to construct a model with favorable invariances, showing that this can even be achieved with the variances present within even a single domain.
Central to our paper is a consideration of the biases which enables a model to generalize.
We show by correcting this common bias in existing CNNs, models can be made to improve performance in general settings without the need for domain specific data.
\section*{Acknowledgment}
This work was funded by DARPA's Learning with Less Labels (LwLL) program under agreement HR0011-18-S-0044.
\clearpage

\section*{Appendix}
\appendix
\section{Detailed Experimental Results}

Tables \ref{PACS-table-std-resnet}, \ref{PACS-table-std-alexnet}, \ref{VLCS-table-std}, \ref{OH-table-std} contain the same results reported in Section \ref{sec:experiments} with standard deviations added on. All experiments are reported as 3 run averages with standard deviations, run on a mix of machines which typically have a 28-core CPU, 384GB RAM with Nvidia RTX 2080Ti, Titan X and Titan Xp gpus. On such machines a results for a 3 run average for the PACS or VLCS dataset takes 4-6 hours while the OH dataset takes 6-8 hours.

\section{Ablation of Style Probability}

Table \ref{style_p_abl} includes an ablation of style probabilities and the effect it has on PACS accuracy. It is clear that there is an amount of stylization that is optimal and stylization beyond this results in reduced performance. However, the results very stable across large changes in probabilities (e.g. 0.25 and 0.50) showing that the method is not overly sensitive to this hyper-parameter.

\begin{table}[h]
  \caption{Prediction accuracy of Stylized training on the PACS Dataset with ResNet-18 as compared to recent state-of-the-art methods. Each reported accuracy is the average of three independent runs of the method with standard deviations added. The runs with an asterisk(*) denote a different training/testing procedure which computes max target dataset accuracy over the training period in lieu of using in source validation data for model selection.}
  \label{PACS-table-std-resnet}
  \centering
  \resizebox{\textwidth}{!}{
  \begin{tabular}{lllll|l}
    \toprule
     & \multicolumn{4}{c}{ResNet-18 - PACS} \\
    \midrule
    Method  & Photo & Art   & Cartoon   & Sketch    & Avg. \\
    \midrule
    D-SAM~\cite{d2019domain} & 95.30 & 77.33 & 72.43 & 77.83 & 80.72 \\
    MMLD~\cite{matsuura2019domain} & 96.09 & 81.28 & 77.16 & 72.29 & 81.83 \\
    Meta-Reg~\cite{balaji2018metareg} & 95.50 $\pm$ 00.24 & 83.70 $\pm$ 00.19 & 77.20 $\pm$ 00.31 & 70.30 $\pm$ 00.28 & 81.70 \\
    Epi-FCR~\cite{li2019episodic} & 93.90 & 82.10 & 77.00 & 73.00 & 81.50 \\
    JiGen~\cite{carlucci2019domain} & 96.03 & 79.42 & 75.25 & 71.35 & 80.51 \\
    Ours~(Painting) & \bf{96.29} $\pm$ 00.03 & \bf{84.60} $\pm$ 1.22 & 78.81 $\pm$ 00.17 & 78.90 $\pm$ 00.84 & \bf{84.64} $\pm$ 00.14 \\
    Ours~(Inter-Source) & 96.27 $\pm$ 00.19 & 83.17 $\pm$ 00.61 & \bf{78.90} $\pm$ 00.15 & \bf{79.87} $\pm$ 00.60 & 84.55 $\pm$ 00.15 \\
    \midrule
    MASF~\cite{dou2019domain}* & 94.99 $\pm$ 0.09 & 80.29 $\pm$ 0.18 & 77.17 $\pm$ 0.08  & 71.69 $\pm$ 0.22 & 81.79 \\
    Ours~(Painting)* & 96.85 $\pm$ 00.19 & \bf{85.45} $\pm$ 00.29 & 79.56 $\pm$ 00.46 & 79.32 $\pm$ 00.17 & 85.30 $\pm$ 00.08 \\
    Ours~(Inter-Source)* & \bf{96.89} $\pm$ 00.45 & 84.33 $\pm$ 00.55 & \bf{80.15} $\pm$ 00.30 & \bf{81.53} $\pm$ 00.47 & \bf{85.72} $\pm$ 00.40 \\
    \bottomrule
  \end{tabular}
  }
\end{table}

\begin{table}[h]
  \caption{Prediction accuracy of Stylized training on the PACS Dataset with AlexNet as compared to recent state-of-the-art methods. Each reported accuracy is the average of three independent runs of the method with standard deviations added. The runs with an asterisk(*) denote a different training/testing procedure which computes max target dataset accuracy over the training period in lieu of using in source validation data for model selection.}
  \label{PACS-table-std-alexnet}
  \centering
  \resizebox{\textwidth}{!}{
  \begin{tabular}{lllll|l}
    \toprule
     & \multicolumn{4}{c}{AlexNet - PACS} \\
    \midrule
    Method              & Photo & Art   & Cartoon   & Sketch    & Avg.\\
    \midrule
    D-SAM~\cite{d2019domain}          & 85.55 & 63.87 & 70.70 & 66.66 & 71.20 \\
    MMLD~\cite{matsuura2019domain}    & 89.00 & 69.27 & \bf{72.83} & 66.44 & 73.38\\
    MLDG~\cite{li2018learning}        & 88.00 & 66.23 & 66.88 & 58.96 & 70.01 \\
    Meta-Reg~\cite{balaji2018metareg} & \bf{91.07} $\pm$ 00.41 & 69.82 $\pm$ 00.76 & 70.35 $\pm$ 00.63 & 59.26 $\pm$ 00.31 & 72.62\\
    Epi-FCR~\cite{li2019episodic}     & 86.10 & 64.70 & 72.30 & 65.00 & 72.00\\
    JiGen~\cite{carlucci2019domain}   & 88.98 & 67.63 & 71.71 & 65.18 & 74.38\\
    PAR~\cite{wang2019learning}       & 90.40 & 68.70 & 70.50 & 64.60 & 73.54\\
    Ours~(Painting)     & 90.00 $\pm$ 00.75 & \bf{73.71} $\pm$ 01.68 & 72.37 $\pm$ 00.44 & 70.40 $\pm$ 02.28 & \bf{76.62} $\pm$ 01.19 \\
    Ours~(Inter-Source) & 89.76 $\pm$ 00.36 & 72.52 $\pm$ 00.84 & 71.72 $\pm$ 02.31 & \bf{71.21} $\pm$ 01.56 & 76.30 $\pm$ 01.08 \\
    \midrule
    MASF~\cite{dou2019domain}* & 90.68 $\pm$ 0.12 & 70.35 $\pm$ 0.33 & 72.46 $\pm$ 0.19 & 67.33 $\pm$ 0.12 & 75.21\\
    Ours~(Painting)*        & \bf{91.60} $\pm$ 00.58 & \bf{76.01} $\pm$ 00.35 & \bf{73.82} $\pm$ 00.54 & 73.07 $\pm$ 01.13 & \bf{78.62} $\pm$ 00.19 \\
    Ours~(Inter-Source)*    & 91.52 $\pm$ 00.24 & 74.59 $\pm$ 00.83 & 73.15 $\pm$ 00.15 & \bf{74.71} $\pm$ 00.94 & 78.49 $\pm$ 00.26 \\
    \bottomrule
  \end{tabular}
  }
\end{table}
\begin{table}[h]
  \caption{Prediction Accuracy of Stylized training on the VLCS Dataset as compared to recent state of the art methods. We report average accuracy and standard deviation over 3 independent runs of the method. The runs with an asterisk(*) denote a different training/testing procedure which computes max target dataset accuracy over the training period in lieu of using in source validation data for model selection.}
  \label{VLCS-table-std}
  \centering
  \resizebox{\textwidth}{!}{
  \begin{tabular}{lllll|l}
    \toprule
    \multicolumn{6}{c}{AlexNet - VLCS} \\
    \midrule
                        & Pascal(VOC)   & LabelMe   & Caltech   & Sun    & Avg. \\
    \midrule
    MMD-AAE~\cite{li2018learning}             & 67.70         & 62.60     & 94.90     & 64.90                      & 72.28 \\
    D-SAM~\cite{d2019domain}               & 58.59         & 56.95     & 91.75     & 60.84                      & 67.03 \\
    MMLD~\cite{matsuura2019domain}                & \bf{71.96}         & 58.77     & 96.66     & 68.13                      & \bf{73.88} \\
    Epi-FCR~\cite{li2019episodic}             & 67.10         & \bf{64.30}     & 94.10     & 65.90                      & 72.90 \\
    JiGen~\cite{carlucci2019domain}               & 70.62         & 60.90     & \bf{96.93}     & 64.30                      & 73.19 \\
    Ours~(Painting)     & 69.81 $\pm$ 00.88        & 58.90 $\pm$ 01.16      & 96.57 $\pm$ 00.46     & \bf{69.82} $\pm$ 00.48                    & 73.77 $\pm$ 00.24 \\
    Ours~(Inter-Source) & 71.17 $\pm$ 01.49 & 60.19 $\pm$ 00.71 & 96.27 $\pm$ 00.87 & 67.68 $\pm$ 01.68                      & 73.83 $\pm$ 00.77 \\
    \midrule
    MASF~\cite{dou2019domain}*                   & 69.14 $\pm$ 00.19 & \bf{64.90} $\pm$ 00.08 & 94.78 $\pm$ 00.16 & 67.64 $\pm$ 00.12    & 74.11\\
    Ours~(Painting)*        & 72.04 $\pm$ 00.90 & 61.97 $\pm$ 00.65 & \bf{97.81} $\pm$ 00.12 & 70.67 $\pm$ 00.26    & 75.62 $\pm$ 00.26\\
    Ours~(Inter-Source)*    & \bf{72.30} $\pm$ 00.42 & 61.95 $\pm$ 00.08 & 97.51 $\pm$ 00.41 & \bf{70.93} $\pm$ 00.37     & \bf{75.67} $\pm$ 00.24\\
    \bottomrule
  \end{tabular}
  }
\end{table}
\begin{table}[h]
  \caption{Prediction Accuracy of Stylized training on the OH Dataset as compared to recent state of the art methods. We report average accuracy and standard deviation over 3 independent runs of the method.}
  \label{OH-table-std}
  \centering
  \resizebox{\textwidth}{!}{
  \begin{tabular}{lllll|l}
    \toprule
    \multicolumn{6}{c}{ResNet-18 - OfficeHome} \\
    \midrule
                        & Art & Clipart & Product   & Real    & Avg. \\
    \midrule
    D-SAM~\cite{d2019domain}               & 58.03 & 44.37 & 69.22     & 71.45     & 60.77\\
    JiGen~\cite{carlucci2019domain}               & 53.04 & 47.51 & 71.47     & 72.79     & 61.20\\
    Ours~(Painting)     & \bf{59.87} $\pm$ 00.50 & 49.91 $\pm$ 00.31 & \bf{72.82} $\pm$ 00.26     & \bf{75.20} $\pm$ 00.15     & \bf{64.45} $\pm$ 00.13\\
    Ours~(Inter-Source) & 59.52 $\pm$ 00.45 & \bf{50.86} $\pm$ 00.20 & 72.15 $\pm$ 00.12    & 75.13 $\pm$ 00.23     & 64.42 $\pm$ 00.21\\
    \bottomrule
  \end{tabular}
  }
\end{table}
\begin{table}[h!]
  \caption{Style Probability vs. PACS Accuracy}
  \label{style_p_abl}
  \centering
  \begin{tabular}{lllll|l}
    \toprule
    \multicolumn{6}{c}{ResNet18 - PACS} \\
    \midrule
    Style Prob. & Photo & Art & Cartoon & Sketch & Avg \\
    \midrule
    0.25  & 96.07 & 84.03 & 77.57 & 79.68 & 84.34 \\
    0.50  & 95.57 & 83.41 & 77.05 & 80.40 & 84.11 \\
    0.75  & 95.19 & 83.06 & 75.23 & 79.90 & 83.37 \\
    1.00  & 94.71 & 82.79 & 72.13 & 75.36 & 81.25 \\
    \bottomrule
  \end{tabular}
\end{table}

\section{Stylization Training Comparison}

\begin{figure}[h!]
    \begin{center}
    \includegraphics[width=0.85\linewidth]{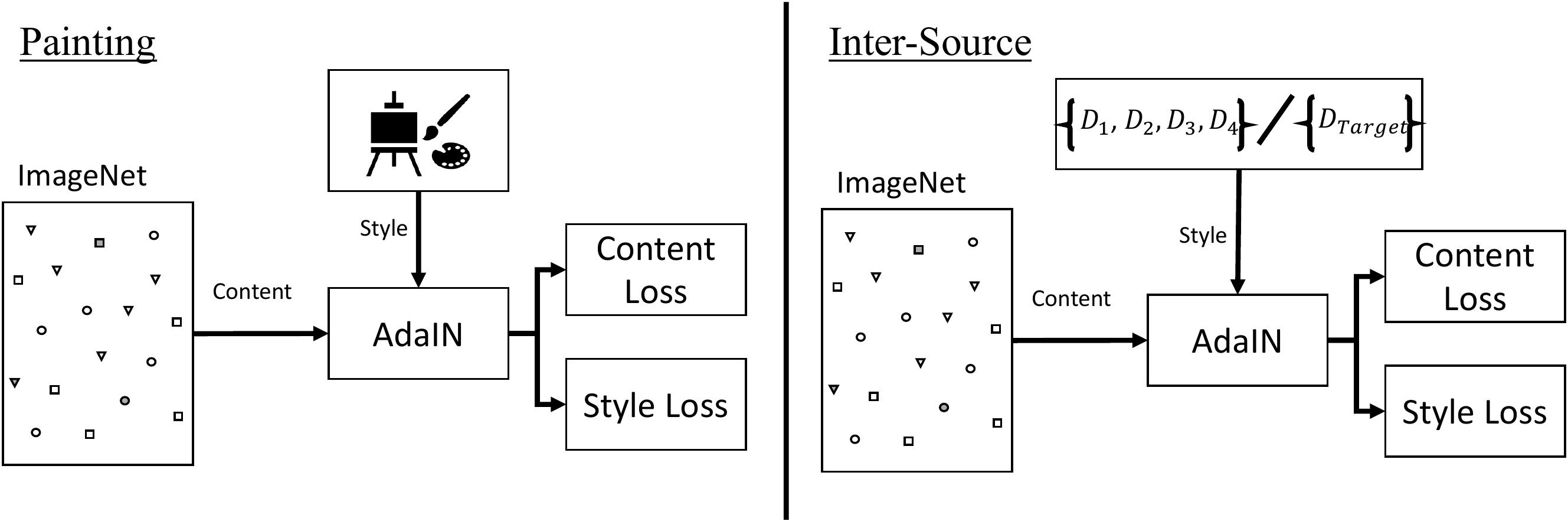}
    \caption{
        Illustration of training pipeline for retraining AdaIN in both the painting stylization and inter-source setting.
    }
    \label{fig:adain_retrained}
    \end{center}
\end{figure}

We explore different methods of training the AdaIN stylization network in our two stylization methods. 
The original AdaIN network is trained with the COCO dataset for content images and the Painter By Numbers dataset as style. 
First, we explore alternatively training the network with ImageNet images as content and Painter By Numbers as the style data for the painting stylization setting. 
Second, we explore re-training the network in the inter-source setting by again using ImageNet images as content images, but using the relevant DG dataset (with target removed) as the style data.
This results in one stylization model for stylizing images as paintings and 12 different stylization models for inter-source stylization, one for each target domain in each DG dataset.
For example, an AdaIN model trained to be used for inter-source training on the PACS dataset with  \textit{Sketch} as the target domain would be trained with data from the \textit{Photo}, \textit{Art} and \textit{Cartoon} datasets as style data.
Figure \ref{fig:adain_retrained} depicts this visually. 
Note that we did not perform any tuning for this training, which may result in further performance improvements. 
This is due to the fact that tuning this method would require runs of stylization and model training for just a single data point, making it prohibitively expensive.
The results of this experiment are in Tables~\ref{retrained-PBN} and \ref{retrained-inter-source}. In each setting retraining AdaIN results in similar performance indicating that the stylization method is robust the specific choice of datasets used.

\begin{table}[t]
  \caption{Results of retraining the AdaIN stylization model in the \textbf{painting stylization setting} compared to stylization via the pretrained AdaIN model. In each case the mean and standard deviation is reported over three independent runs.}
  \label{retrained-PBN}
  \centering
  \resizebox{\textwidth}{!}{
  \begin{tabular}{lllll|l}
    \toprule
    \multicolumn{6}{c}{ResNet18} \\
    \midrule
    Method  & Photo & Art   & Cartoon   & Sketch    & Avg. \\
    \midrule
    PACS~(Pre-Trained) & \bf{96.29} $\pm$ 0.03 & \bf{84.60} $\pm$ 1.22 & 78.81 $\pm$ 0.17 & \bf{78.90} $\pm$ 0.84 & \bf{84.64} $\pm$ 0.14 \\
    PACS~(Re-Trained) & \bf{96.23} $\pm$ 0.31 & 83.35 $\pm$ 0.81 & \bf{79.74} $\pm$ 0.38 & 77.46 $\pm$ 0.13 & 84.19 $\pm$ 0.22 \\
    \midrule
    Method  & Art & Clipart   & Product   & Real    & Avg. \\
    \midrule
    OH~(Pre-Trained) & \bf{59.87} $\pm$ 0.50 & \bf{49.91} $\pm$ 0.31 & \bf{72.82} $\pm$ 0.26 & \bf{75.20} $\pm$ 0.15 & \bf{64.45} $\pm$ 0.13\\
    OH~(Re-Trained) & \bf{59.62} $\pm$ 0.22 & \bf{49.65} $\pm$ 0.39 & 72.25 $\pm$ 0.06 & \bf{75.08} $\pm$ 0.23 & 64.15 $\pm$ 0.10 \\
    \midrule
    \multicolumn{6}{c}{AlexNet} \\
    \midrule
    Method  & Photo & Art   & Cartoon   & Sketch    & Avg. \\
    \midrule
    PACS~(Pre-Trained) & \bf{90.00} $\pm$ 0.75 & \bf{73.71} $\pm$ 1.68 & 72.37 $\pm$ 0.44 & 70.40 $\pm$ 2.28 & \bf{76.62} $\pm$ 1.19 \\
    PACS~(Re-Trained) & \bf{90.04} $\pm$ 0.09 & \bf{72.25} $\pm$ 1.78 & \bf{70.80} $\pm$ 0.28 & \bf{69.19} $\pm$ 0.79 & \bf{75.57} $\pm$ 0.58 \\
    \midrule
    Method  & Pascal (VOC) & LabelMe   & Caltech   & Sun    & Avg. \\
    \midrule
    VLCS~(Pre-Trained) & \bf{69.81} $\pm$ 0.88 & \bf{58.90} $\pm$ 1.16 & \bf{96.57} $\pm$ 0.46 & \bf{69.82} $\pm$ 0.48 & \bf{73.77} $\pm$ 0.24  \\
    VLCS~(Re-Trained) & \bf{69.91} $\pm$ 0.34 & \bf{59.48} $\pm$ 1.74 & \bf{96.54} $\pm$ 0.59 & 66.87 $\pm$ 0.89 & \bf{73.20} $\pm$ 0.58 \\
    \bottomrule
  \end{tabular}
  }
\end{table}

\begin{table}[t]
  \caption{Results of retraining the AdaIN stylization model in the \textbf{inter-source stylization setting} compared to stylization via the pretrained AdaIN model. In each case the mean and standard deviation is reported over three independent runs.}
  \label{retrained-inter-source}
  \centering
  \resizebox{\textwidth}{!}{
  \begin{tabular}{lllll|l}
   \toprule
    \multicolumn{6}{c}{ResNet18} \\
    \midrule
    Method  & Photo & Art   & Cartoon   & Sketch    & Avg. \\
    \midrule
    PACS~(Pre-Trained) & \textbf{96.27 $\pm$ 0.19} & \textbf{83.17 $\pm$ 0.61} & 78.90 $\pm$ 0.15 & 79.87 $\pm$ 0.60 & \textbf{84.55 $\pm$ 0.15} \\
    PACS~(Re-Trained) & \textbf{96.43 $\pm$ 0.15} & 80.31 $\pm$ 0.23 & \textbf{79.74 $\pm$ 0.45} & \textbf{77.55 $\pm$ 1.04} & 83.51 $\pm$ 0.31 \\
    \midrule
    Method  & Art & Clipart   & Product   & Real    & Avg. \\
    \midrule
    OH~(Pre-Trained) & \textbf{59.52 $\pm$ 0.45} & \textbf{50.86 $\pm$ 0.20} & \textbf{72.15 $\pm$ 0.12} & \textbf{75.13 $\pm$ 0.23}     & \textbf{64.42 $\pm$ 0.21}\\
    OH~(Re-Trained) & \textbf{60.06 $\pm$ 0.19} & \textbf{51.35 $\pm$ 0.51} & \textbf{72.29 $\pm$ 0.18} & 74.68 $\pm$ 0.20 & \textbf{64.60 $\pm$ 0.17} \\
    \midrule
    \multicolumn{6}{c}{AlexNet} \\
    \midrule
    Method  & Photo & Art   & Cartoon   & Sketch    & Avg. \\
    \midrule
    PACS~(Pre-Trained) & \textbf{89.76 $\pm$ 0.36} & \textbf{72.52 $\pm$ 0.84} & \textbf{71.72 $\pm$ 2.31} & \textbf{71.21 $\pm$ 1.56} & \textbf{76.30 $\pm$ 1.08} \\
    PACS~(Re-Trained) & \textbf{90.08 $\pm$ 0.33} & \textbf{71.84 $\pm$ 1.92} & 70.45 $\pm$ 0.21 & \textbf{70.37 $\pm$ 0.75} & \textbf{75.68 $\pm$ 0.45} \\
    \midrule
    Method  & Pascal (VOC) & LabelMe   & Caltech   & Sun    & Avg. \\
    \midrule
    VLCS~(Pre-Trained) & \textbf{71.17 $\pm$ 1.49} & \textbf{60.19 $\pm$ 0.71} & \textbf{96.27 $\pm$ 0.87} & \textbf{67.68 $\pm$ 1.68} & \textbf{73.83 $\pm$ 0.77} \\
    VLCS~(Re-Trained) & \textbf{71.18 $\pm$ 0.66} & 58.06 $\pm$ 1.24 & \textbf{96.27 $\pm$ 1.05} & \textbf{68.02 $\pm$ 1.19} & \textbf{73.38 $\pm$ 0.66} \\
    \bottomrule
  \end{tabular}
  }
\end{table}

\clearpage
\bibliographystyle{abbrvnat}
\bibliography{reference}

\end{document}